\documentclass{article}
\usepackage{ijcai24}

\usepackage{times}
\usepackage{soul}
\usepackage{url}
\usepackage[hidelinks]{hyperref}
\usepackage[utf8]{inputenc}
\usepackage[small]{caption}
\usepackage{graphicx}
\usepackage{amsmath}
\usepackage{amsthm}
\usepackage{booktabs}
\usepackage{algorithm}
\usepackage{algorithmic}
\usepackage[switch]{lineno}

\usepackage{bm}
\usepackage{amssymb}
\usepackage{amsfonts}
\usepackage{subfigure}
\usepackage{threeparttable}
\usepackage{tablefootnote}
\usepackage{colortbl}
\usepackage[dvipsnames]{xcolor}
\usepackage{wasysym}
\usepackage{multirow}
\usepackage{appendix}
\usepackage[T1]{fontenc}
\usepackage{nopageno}

\title{Reassessing Evaluation Functions in Algorithmic Recourse:\\An Empirical Study from a Human-Centered Perspective}

\author{
Tomu Tominaga$^1$\and
Naomi Yamashita$^2$\and
Takeshi Kurashima$^1$\\
\affiliations
$^1$NTT Human Informatics Laboratories, NTT Corporation\\
$^2$NTT Communication Science Laboratories, NTT Corporation\\
\emails
\{tomu.tominaga, takeshi.kurashima\}@ntt.com,
naomiy@acm.org
}

\begin{document}

\maketitle

\begin{abstract}
In this study, we critically examine the foundational premise of {\it algorithmic recourse} -- a process of generating counterfactual action plans (i.e., recourses) assisting individuals to reverse adverse decisions made by AI systems.
The assumption underlying algorithmic recourse is that individuals accept and act on recourses that minimize the gap between their current and desired states.
This assumption, however, remains empirically unverified.
To address this issue, we conducted a user study with 362 participants and assessed whether minimizing the distance function, a metric of the gap between the current and desired states, indeed prompts them to accept and act upon suggested recourses.
Our findings reveal a nuanced landscape: participants' acceptance of recourses did not correlate with the recourse distance.
Moreover, participants' willingness to act upon recourses peaked at the minimal recourse distance but was otherwise constant.
These findings cast doubt on the prevailing assumption of algorithmic recourse research and signal the need to rethink the evaluation functions to pave the way for human-centered recourse generation.
\end{abstract}

\section{Introduction}
As artificial intelligence (AI) has played an active role in high-stake decision making, algorithmic recourse has gained significant attention as a key explainable AI (XAI) technology.
To help individuals who receive unfavorable decisions from AI systems, algorithmic recourse provides a counterfactual action plan, called \textit{recourse}, for them to flip the decision and reach a favorable outcome~\cite{Karimi2022} such as ``your loan application would be approved if you increased the annual income by \$10K and changed the educational background from bachelor to master''.
Its ultimate goal is to offer recourses for unfavorable decisions so that individuals can accept and act upon them.

To generate such optimal recourses, algorithmic recourse research has addressed the challenge of identifying the counterfactual sample that has minimal disparity with the target individual.
Here, counterfactual samples refer to user samples chosen to contrast with target individuals within recourses, selected from among those who receive favorable decisions from the AI system.
The disparity between target and counterfactual individuals is often evaluated with distance functions that quantify the feature differences between them using norm-based measures~\cite{Verma2020}.
This task is typically formulated as an optimization problem~\cite{Wachter2018}, with numerous recent proposals for technical solutions~\cite{Karimi2022,Verma2020}.

Such technological advancements rest upon a crucial assumption that individuals accept and act on recourses generated through the minimization of the evaluation functions.
To date, the evaluation functions have been developed to capture the simplicity of explanations and the effortlessness of action plans in the suggested recourses.
Concerning simplicity, researchers presume that individuals prefer recourses with fewer changes suggested.
This is inspired by Miller's insights~\cite{Miller2019}, emphasizing people's preference for simpler explanations with fewer causes cited~\cite{Thagard1989,Read1993}.
For effortlessness, acknowledging that transitioning from the current to the desired state incurs costs, they posit that recourses with minimal suggested changes are easier for individuals to implement~\cite{Ustun2019,Karimi2022}.
Hence, it has been presumed that recourses minimizing proposed changes assessed by evaluation functions are optimal for individuals.

However, little empirical evidence supports the underlying assumption that individuals indeed accept and act upon the recourses minimizing the evaluation functions.
As evidenced in persuasive technology research~\cite{Fogg1998}, reshaping human attitudes and behaviors through computational approaches is challenging, suggesting that the recourse generation process is more nuanced than initially assumed.
This lack of validation for the aforementioned assumption can be attributed to the XAI community's focus on technological advancements, often overlooking the importance of rigorous evaluation through user studies~\cite{Keane2021,Adadi2018}.
To achieve robust scientific progress towards human-centered algorithmic recourse generation, it is imperative to confirm the validity of the evaluation function -- the core principle of recourse generation that acts as the objective function in the optimization problem. 
This validation should be achieved through user studies~\cite{Barocas2020} and psychologically grounded~\cite{Keane2021}.

In this paper, we aim to provide empirical evidence for the premise underlying algorithmic recourse by addressing the following research question.
\begin{description}    
\item[Research Question:] \textit{How does the recourse distance, as quantified by evaluation functions, affect individuals' willingness to accept and act upon recourses?}
\end{description}
This study employs $L_0$ and $L_1$ norms, referred to as \textit{sparsity} and \textit{proximity} respectively, as the evaluation functions in this study due to their fundamental roles in algorithmic recourse research~\cite{Karimi2022,Verma2020}.

To investigate the research question, we conducted an online study with 362 participants.
Our experiment involved a real-world car loan application scenario, as financial assessments such as loan screening or credit evaluation are one of the paramount themes in algorithmic recourse research~\cite{Kirfel2021,Wang2023}.
To enhance the experiment's authenticity, we recruited participants genuinely interested in a car loan.
We systematically devised various recourses tailored to participants' profiles, controlling the recourse distance using the evaluation functions, and evaluated their inclinations to accept and act upon these recourses.

\paragraph{Contributions.}
This study presents a unique contribution by validating and challenging the foundational assumption of algorithmic recourse research through user experiments. 
Notably, it provides empirical evidence that participants' willingness to accept recourses was not affected by the recourse distance.
Furthermore, the likelihood of acting on recourses was maximized at minimal recourse distance but remained constant elsewhere.
These findings underscore the urgent need to revisit the evaluation functions used in algorithmic recourse.
Additionally, this study enriches the field by outlining future research directions crucial for developing a human-centered approach to algorithmic recourse generation AI.

\section{Related Work}
\subsection{Algorithmic Recourse Generation}
In the rapidly evolving research field, the definition of algorithmic recourse varies~\cite{Joshi2019,Ustun2019,Venkatasubramanian2020}, but its overarching objective is consistent: to help individuals subjected to unfavorable AI decisions understand the rationale behind them~\cite{Joshi2019,Wachter2018} and subsequently achieve favorable outcomes~\cite{Karimi2021,Karimi2020}.
Drawing on the concept that individuals favor simpler explanations~\cite{Miller2019} and are more inclined to undertake recourse that proposes smaller changes~\cite{Karimi2022}, researchers have tackled the following optimization problem to identify counterfactual samples for constructing recourses: given a fixed predictive model $h:X\rightarrow\{-1,+1\}$ (e.g., $-1$ is ``rejection'' and $+1$ is ``approval'') with an $N$-dimension feature space $X=X_1\times \cdots \times X_N$ and its subspace $S(X)\subseteq X$ determined by arbitrary conditions on the features, find a counterfactual sample $x'\in X$ of an input sample $x\in X$ ($h(\bm{x})=-1$).
\begin{equation}
    \bm{x'} \in \underset{\bm{x'}\in S(X)} {\operatorname{argmin}}\ d(\bm{x}, \bm{x'})\ \text{subject to}\ h(\bm{x'})\neq h(\bm{x})\\
\end{equation}
In this task, $d(\cdot,\cdot):X\times X\rightarrow \mathbb{R}_{\geq0}$ is a distance function to measure the distance between $\bm{x}$ and $\bm{x'}$.
Among various types of distance functions, the most fundamental ones are norm-based measures (e.g., $L_1$ norm)~\cite{Verma2020}.

To date, many studies have predominantly focused on enhancing technical performance such as capturing feature distributions~\cite{Kanamori2020,Poyiadzi2020}, incorporating causal knowledge among features into constraints~\cite{Karimi2021,Karimi2020a}, making coherent and diverse explanations~\cite{Russell2019,Mothilal2020}, or computing the order of actions~\cite{Kanamori2021,Naumann2021}.
While the optimization problem inherently suggests that minimizing the distance functions will lead algorithmic recourse to produce recourses that are both acceptable and actionable for individuals, empirical evidence supporting this notion is scarce.
This study aims to address this issue by conducting user experiments to verify the aforementioned assumption.

\subsection{Human Evaluation of Algorithmic Recourse}
Since the lack of human-centered evaluation for counterfactual suggestions was identified~\cite{Keane2021,Verma2020}, there has been a gradual increase in user studies~\cite{Forster2020,Forster2021,Kanamori2022,Kirfel2021,Rawal2020,Singh2023,Warren2023,Yacoby2022}; however, studies investigating individuals' willingness to accept and act on counterfactual suggestions remain limited.
Among the limited research, an exploratory study using an interactive recourse generation system has shown that recourses with fewer and more controllable changes were actionable~\cite{Wang2023}. 
Additionally, players in a simulation game were found to adopt strategies that mirrored their previously employed tactics~\cite{Kuhl2022}.

However, these studies do not use evaluators' profile data to generate recourses~\cite{Wang2023,Kirfel2021} and ignore real-world contexts~\cite{Kuhl2022}, thereby failing to accurately replicate realistic scenarios where people encounter adverse AI decisions. 
Moreover, these studies focus more on generating recourses rather than evaluating users' preferences. 
Our study aims to bridge this gap by constructing recourses based on participants' profile data within real-world contexts, offering a more authentic evaluation. 
This approach not only confirms the validity of evaluation functions but also enhances our understanding of users' propensity to accept and act on recourses.

\section{Experiment}
We carried out an experiment to assess how individuals' attitudes towards accepting and acting on recourses change with the recourse distance.
To simulate real-world scenarios where individuals encounter unfavorable AI decisions, we developed a scenario involving car loan applications and selectively recruited participants who fit this specific context.
In this scenario, participants submitted their profile information for the application, received a rejection notification based on the assessment, and then rated their willingness to accept and act upon the suggested recourses.

\subsection{Experimental Setup}
\subsubsection{Hypothetical Scenario for Recourse Provision}
We crafted a hypothetical scenario where participants were asked to imagine themselves applying for a car loan at a financial institution.
There are two reasons for choosing the scene of car loans.
First, finance is one of the most promising domains in algorithmic recourse research~\cite{Ustun2019,Karimi2020,Wang2023}, with loan or credit screening being a predominant case~\cite{Kirfel2021,Wang2023}.
Second, among items financed through loans such as education loans, car loans, and mortgages, cars are typically well-known, relatively expensive, and are actively owned and used by the buyer; therefore, the car loan application scenario is likely to prompt participants to easily imagine themselves in the situation, think it as a high-stakes decision, and relate it to their real-life context when evaluating recourses.

\subsubsection{Scenario Design}
The participants engaged in the following car loan application scenario:
\begin{quote}
  \textit{
  The participant wants to take out a two-year car loan equivalent to one-third of their annual income. 
  They visit a financial institution to obtain the loan and are asked to submit their profile data for screening. 
  The screening process involves an AI system that determines applicants' eligibility based on an extensive database of customer information held by the financial institution. 
  After submitting their profile data, they are informed of the rejection of their loan application shortly thereafter. 
  To find out why the application was rejected and what action is required to secure approval, they will view the recourses generated by the AI system, derived from their profile data and the financial institution's customer database.}
\end{quote}

Here, the loan amount in this experiment is relative to the participants' annual income rather than a fixed amount, with a repayment term fixed at two years.
This scenario design ensures uniformity in the repayment burden across participants, thereby minimizing its influence on their evaluation of the likelihood of acting upon the suggested recourses.

We constructed recourses based on the assumption that the financial institution has a fixed rule of rejecting any loan applications exceeding one-quarter of the applicant's annual income. 
Therefore, it should be noted that all the participants experienced rejection in this experiment.
This rule was not disclosed to the participants throughout the experiment.

\subsection{Recourse Construction~\label{subsec:recourse}}
To enable participants to evaluate recourses across diverse distances, we selected five counterfactual samples for each input sample (i.e., a participant's profile data).
To accomplish this, we pre-collected 4057 profile data points to serve the counterfactual samples, representing ``an extensive database of customer information held by the financial institution'' in the scenario (see Section A in supplementary materials for further details).
Using this counterfactual sample pool, we constructed recourses as follows.

\subsubsection{Conditions}
A counterfactual sample $\bm{x'}$, used for generating a recourse $\bm{\delta}$ given an input sample $\bm{x}$, must meet the following conditions: (1) the annual income of $\bm{x'}$ is 4/3 times greater than that of $\bm{x}$ and (2) $\bm{x'}$ and $\bm{x}$ follow the constraints in Table~\ref{tab:profile}.
\begin{table*}[t]
    \begin{tabular}{clp{1.15\columnwidth}l}
        \toprule
        \# & Item (Feature) & Option & Constraints\\
        \midrule
        1 & Residential prefecture & 1. Tokyo / 2. Other than Tokyo & $x'_1\gtreqqless x_1$\\
        2 & Type of residence & 1. Owned house / 2. Rental housing & $x'_2\gtreqqless x_2$\\
        3 & Educational background & 1. High school / 2. Junior college / 3. University (bachelor) / 4. Graduate school (master) / 5. Graduate school (doctor) & $x'_3\geq x_3$\\
        4 & Workplace & 1. Private company / 2. Public institution & $x'_4\gtreqqless x_4$\\
        5 & Position & 1. Employee / 2. Supervisor / 3. Section Head / 4. Section Chief / 5. Assistant General Manager / 6. Manager / 7. General Manager / 8. Executive Director / 9. Senior Executive Director / 10. President & $x'_5\geq x_5$\\
        6 & Service years & 1. 0-1 year / 2. 1-3 years / 3. 3-5 years / 4. 5-10 years / 5. 10-20 years / 6. 20- years & $x'_6\geq x_6$\\
        7 & Management career & 1. No / 2. 0-1 year / 3. 1-3 years / 4. 3-5 years / 5. 5-10 years / 6. 10-20 years / 7. 20- years & $x'_7\geq x_7$\\
        8 & Working hours per day & 1. 0-2 hours / 2. 2-4 hours / 3. 4-6 hours / 4. 6-8 hours / 5. 8-10 hours / 6. 10-12 hours / 7. 12- hours & $x'_8\gtreqqless x_8$\\
        9 & Teleworking hours per day & 1. 0-2 hours / 2. 2-4 hours / 3. 4-6 hours / 4. 6-8 hours / 5. 8-10 hours / 6. 10-12 hours / 7. 12- hours & $x'_9\gtreqqless x_9$\\
        10 & The number of side jobs & 1. No / 2. 1 job / 3. 2 jobs / 4. 3 jobs / 5. 4 jobs / 6. 5- jobs & $x'_{10}\gtreqqless x_{10}$\\
        11 & Job change experience & 1. No / 2. Yes & $x'_{11}\geq x_{11}$\\
        12 & Overseas working experience & 1. No / 2. Yes & $x'_{12}\geq x_{12}$\\
        13 & Overseas study experience & 1. No / 2. Yes & $x'_{13}\geq x_{13}$\\
        14 & TOEIC$^{\ast}$ Best score & 1. No / 2. 10-400 / 3. 400-495 / 4. 500-595 / 5. 600-695 / 6. 700-795 / 7. 800-895 / 8. 900-990 & $x'_{14}\geq x_{14}$\\
        15 & Facebook app use & 1. No / 2. Yes & $x'_{15}\gtreqqless x_{15}$\\
        16 & LinkedIn app use & 1. No / 2. Yes & $x'_{16}\gtreqqless x_{16}$\\
        \bottomrule 
    \end{tabular}
    \caption{Profile data included in recourses. $x'_i$ and $x_i$ are $i$-th elements of an input and counterfactual samples (e.g., $x'_5=2$ means that the counterfactual sample's Position is a Supervisor). The constrains column describes feature conditions of a counterfactual sample $\bm{x'}$ to be selected for a suggested recourse given an input sample $\bm{x}$. \\$^{\ast}$The Test of English for International Communication (TOEIC®) Listening \& Reading Test (\url{https://www.iibc-global.org/english/toeic/test/lr.html}), one of the most popular English language proficiency tests in Japan. The score is in 5-point increments from 10 to 990.~\label{tab:profile}}
\end{table*}
The first one echoes the fixed decision rule of the financial institution in the experimental scenario.
Under the fixed rule, the car loan applications are accepted if the annual income is 4 times or more of the loan amount.
In the scenario, the loan amount applied is one third of the annual income.
We then set the first condition to select counterfactual samples for recourse customization.

The second condition relates the feature constraints.
As described in Table~\ref{tab:profile}, these constraints aim to exclude impossible or extremely rare feature changes.
Examples include raising the educational degree (\#3), work position (\#5), service years (\#6), and management career (\#7), but it is impossible (extremely rare) to lower them.
It is also impossible to remove one's experience and skills such as the job change experience (\#11), overseas working experience (\#12), overseas study experience (\#13), and best test score (\#14) from oneself.
Other than the above features, we do not impose any constraints.
These constraints enable us to generate recourses changing features within a theoretically modifiable range.

\subsubsection{Distance Metrics}
For distance metrics as the evaluation functions of recourse $\bm{\delta}$, we adopted the $L_0$ norm as the sparsity and the $L_1$ norm as the proximity as follows.
Given input sample $\bm{x}$ and its counterfactual sample $\bm{x'}$, we firstly computed
\begin{equation}
  \delta_i = 
    \begin{cases}
    \mathbb{I}[x'_i\neq x_i] & (\text{if $i$-th feature is categorical})\\
    |x'_i-x_i|/M_i & (\text{otherwise})
    \end{cases}\\
\end{equation}
Here, $\mathbb{I}[\cdot]$ is the indicator function.
$M_i$ is the range of the $i$-th feature change to normalize the distances of all features to scales ranging from 0 to 1.
We then calculated
\begin{equation}
  \text{sparsity} = \sum_i \mathbb{I}[\delta_i \neq 0], \ 
  \text{proximity} = \sum_i \delta_i.
\end{equation}

We adopted these metrics due to their foundational significance and widespread acceptance in the research community~\cite{Verma2020}, despite the presence of several other alternative metrics~\cite{Ustun2019,Wachter2018}.
Note that these two metrics exhibit lower values as the recourse is more sparse or proximate, following the technical definitions in prior research~{\cite{Karimi2022,Verma2020,Wachter2018}}.

\subsubsection{Selection Process}
From the subset of the counterfactual sample pool extracted by the above conditions, we selected five counterfactual samples for each participant for recourse customization and evaluation using the following strategy: (1) select one of the sparsest samples, (2) select one of the most proximate samples, and (3) randomly select three additional samples.

\begin{figure}[t]
  \subfigure[Sparsity]{
    \includegraphics[width=0.45\columnwidth]{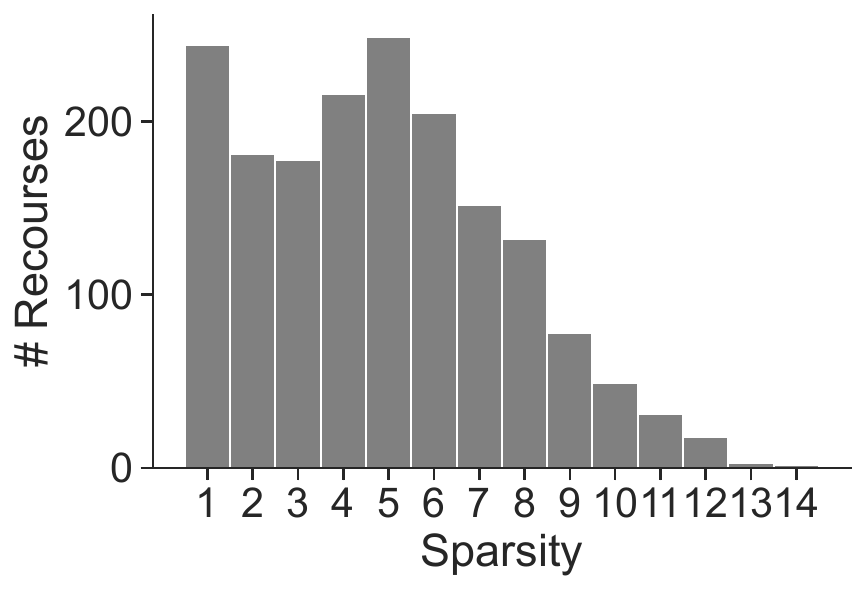}
  }
  \subfigure[Proximity]{
    \includegraphics[width=0.45\columnwidth]{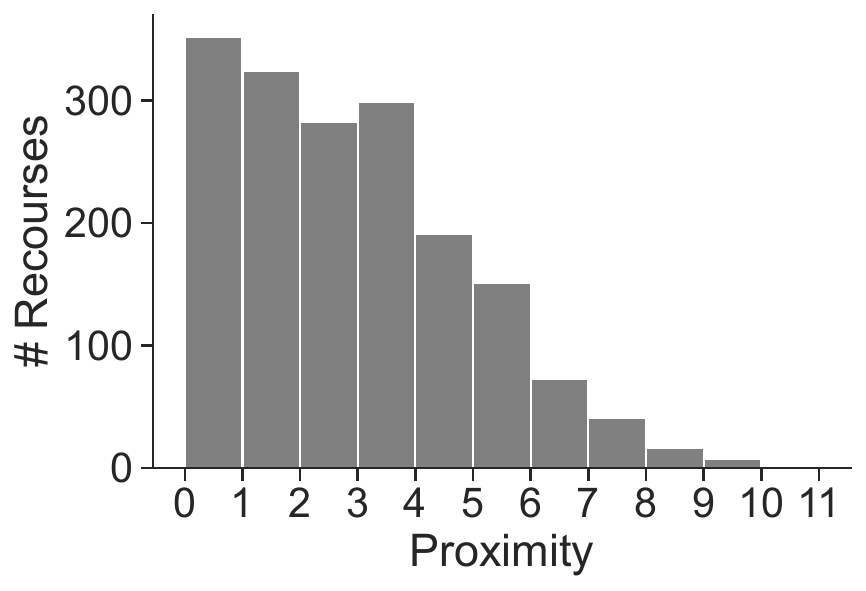}
  }
  \caption{Distributions of distance metrics of selected recourses.\label{fig:distance}}
\end{figure}

Given that the main objective of our experiment is to evaluate participants' responses to recourses across a spectrum of distances, we employed the first and second strategies to select short-distance recourses, while the third strategy was used to introduce a broader variety of distances (Figure~\ref{fig:distance}). 

\subsection{Experimental Procedure}
\begin{figure}[t]
    \centering
    \includegraphics[width=0.95\columnwidth]{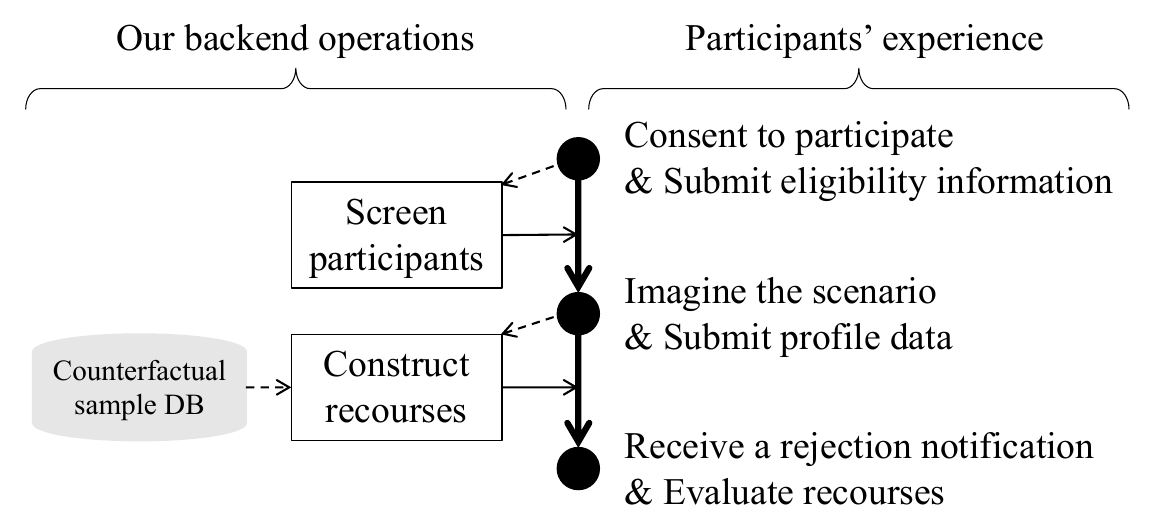}
    \caption{Overview of the experiment process. Black circles indicate steps taken by participants, white boxes represent interventions from us to participants, and thin dashed arrows depict data flow.}
    \label{fig:procedure}
\end{figure}
Participants first took a screening survey to assess their study eligibility, followed by filling out a consent form. 
They then engaged with a hypothetical car loan scenario and submitted their profile data. 
Finally, they received a rejection notification along with five recourses and evaluated each through questionnaires (Figure~\ref{fig:procedure}). 
Details on the questionnaires can be found in Section B of the supplementary materials. 

The experiment was conducted from July 7th to August 10th, 2023.
Participants received a compensation of 100 JPY for their participation.

\subsubsection{Participant Recruitment and Screening}
A total of 362 Japanese participants were recruited through a recruiting company in Japan, meeting the following criteria: (1) employment in private companies or public institutions (full-time or part-time), (2) interest in applying for a car loan, (3) no existing loans, and (4) an annual income less than JPY 10M.
The first criterion targeted paid workers, excluding students, to ensure participants had stable incomes. 
The second and third criteria were designed to select participants interested in and financially capable of considering a car loan.
For instance, those with existing loans might lack the financial flexibility to pursue new ones, leading to reluctance in engaging with new financial commitments.
The fourth criterion facilitated the selection of five counterfactual samples for each participant from the counterfactual sample pool (see Section A in supplementary materials for further details).

\subsubsection{Profile Data Submission~\label{subsubsec:profile}}
Following the screening process, participants provided their profile data relevant to the experimental scenario.
The profile data items are shown in Table~\ref{tab:profile}.
We selected items covering basic demographics (\#1-3), current employment status (\#4-10), professional experience and skills (\#11-14), and human connections (\#15, \#16).
These were chosen to mirror the attributes commonly included in machine learning datasets for financial credit judgments~\cite{GermanCredit1994,TaiwanCredit2016,AdultIncome1996} and information used in a typical resume~\cite{Brown1994,Cole2007}, assuming these reflect critical screening items for evaluating applicants' creditworthiness and employment stability.

Here, we deliberately avoided profile data categorized as immutable features - those that individuals cannot change (e.g., gender, birthplace, or ethnicity)~\cite{Kirfel2021} because recourses including such features sometimes acceptable but always unactionable~\cite{Karimi2022}.

We also excluded annual income from the profile data to maintain the focus of our scenario-based experiment. 
In this experiment, participants' applications were declined due to insufficient annual income relative to the requested loan amount. 
As such, if annual income is included as a profile data item, recourses inevitably advise its adjustment.
This could narrow the focus of our evaluation to participants' reactions to changes in annual income, rather than their perceptions of recourse sparsity. 
To address this and ensure that a diverse range of features are considered in 1-sparsity recourses, annual income is omitted from the profile data.

\subsubsection{Recourse Evaluation}
After the profile data submission, participants were instructed to wait two weeks for the review.
During this period, we developed customized recourses (Section~\ref{subsec:recourse}), refined the questionnaire for evaluating these recourses, and sent it as a review report for each participant.
Upon receiving this report, participants were informed that their applications had been rejected.
They were then asked to evaluate the recourses generated by the AI system as outlined in Table~\ref{tab:recourse}.
\begin{table}[t]
    \centering
    \begin{tabular}{l|c|c|c} \toprule
         & \textcolor{gray}{\frownie{} Current profile} & & \textcolor{ForestGreen}{\smiley{} Ideal profile}\\ \midrule
        $\cdots$ & $\cdots$ & & $\cdots$\\
        Workplace & Private company & & Private company\\
        Position & Employee & $\rightarrow$ & \cellcolor[named]{green} Supervisor\\
        $\cdots$ & $\cdots$ & & $\cdots$\\ \bottomrule
    \end{tabular}
    \caption{A partially excerpted example of a recourse presented to participants (see Section B3 in supplementary materials for a full example). The current profile is the participant's profile and the ideal profile is her/his counterfactual sample. All the items are displayed irrespective of modifications. Items recommended for a change are highlighted in green and indicated with an arrow.\label{tab:recourse}}
\end{table}

To assess the participants' evaluations of recourses, we posed the following two questions: \textit{``Is the AI system's plan a reasonable explanation for the rejection of your loan application?''} to gauge their propensity to accept recourses, and \textit{``Would you carry out the plan presented by the AI system to obtain loan approval?''} to measure their willingness to act upon recourses.
For both questions, responses were captured on a 7-point scale from \textit{``Strongly no''} to \textit{``Strongly yes''}.
Participants were also asked to explain the reasons for their recourse evaluations through an open-ended question.

We also examined the immutability of changing features as suggested in the recourses.
As noted in Section~\ref{subsubsec:profile}, immutable feature changes should be excluded from the recourses because they inherently make recourses unactionable.
Given that the immutability of altering features depends on personal situations, we asked participants about this aspect to filter out impractical recourses from further analyses.

\section{Analysis}
We collected a total of 1810 sets of recourse evaluation data. 
Among them, 72 had at least one immutable feature change. 
After removing these, we analyzed the remaining 1738 evaluations quantitatively and qualitatively. 

\subsection{Quantitative Analysis}
To investigate the connections between individuals' willingness to accept or act on the recourses and recourse distance, we utilized generalized additive mixed models (GAMMs).
This analytical approach combines the characteristics of generalized additive models and mixed-effect models to estimate the nonlinear dependencies between the objective and explanatory variables, accounting for individual subject-specific effects.
We established GAMMs as follows, where $Y$ represents the propensity to accept or act upon recourses, $x$ denotes the recourse distance, and $\epsilon$ denotes the error term, using the parameters $\theta$ and the smoothing spline function $s(\cdot)$.
Note that $i$ denotes the observation, $u$ represents the participant, and $j$ shows the smoothing spline function.
\begin{equation}
    \begin{split}
        Y_{iu} &= \theta_{0u} + \sum^{K}_{j=1}{\theta_{ju}s_j(x_{iu})} + \epsilon_{iu} \\
        \theta_{0u} &= \beta_{00} + b_{0u}, \ b_{0u} \sim \mathcal{N}(0,\sigma_0^2) \\
        \theta_{ju} &= \beta_{j0} + b_{ju}, \ b_{ju} \sim \mathcal{N}(0,\sigma_j^2) 
    \end{split}
\end{equation}
Nesting the random effects $b$ associated with individual participants within the intercept $\theta_0$ and slope $\theta_j$ enables us to discern the fixed effects $\beta$ of recourse distance.
We fitted the aforementioned GAMMs using the mgcv package 1.9-0 421 in RStudio 2023.09.1+494.

\subsection{Qualitative Analysis}
The free-form survey responses added nuance to our quantitative results, shedding light to participants' perspectives on (non-)sparse/proximate recourses, as well as their rationales for favoring one over the other.
Specifically, we examined the responses to the open-ended survey question, \textit{``Why did you make the evaluation you did?''}, following their ratings on propensity for acceptance and action, respectively.

\section{Results}
\begin{figure*}[t]
    \centering
    \subfigure[Smoothing spline curves of sparsity (left) and proximity (right)]{
        \includegraphics[width=0.95\columnwidth]{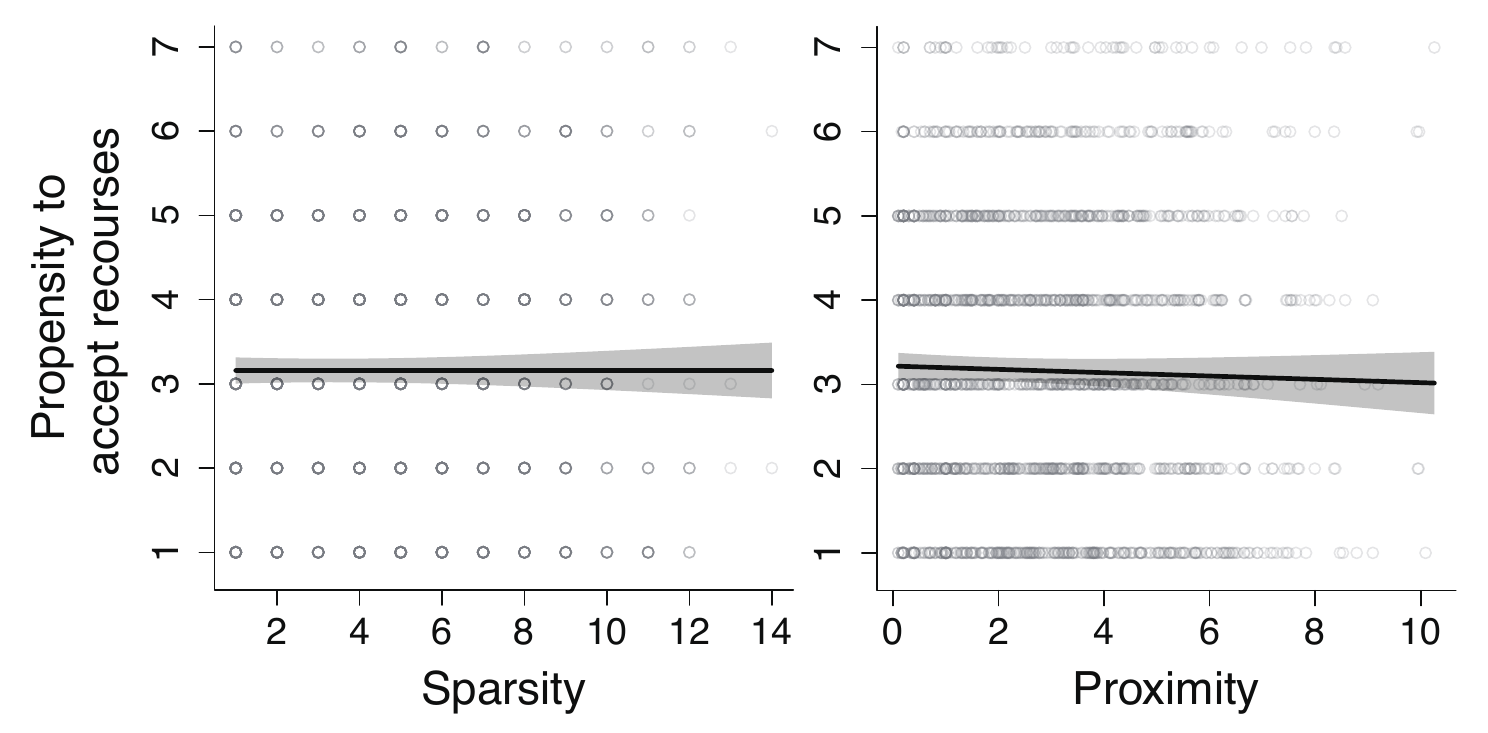}\label{subfig:acp}
    }
    \hspace{2mm}
    \subfigure[Smoothing spline curves of sparsity (left) and proximity (right)]{
        \includegraphics[width=0.95\columnwidth]{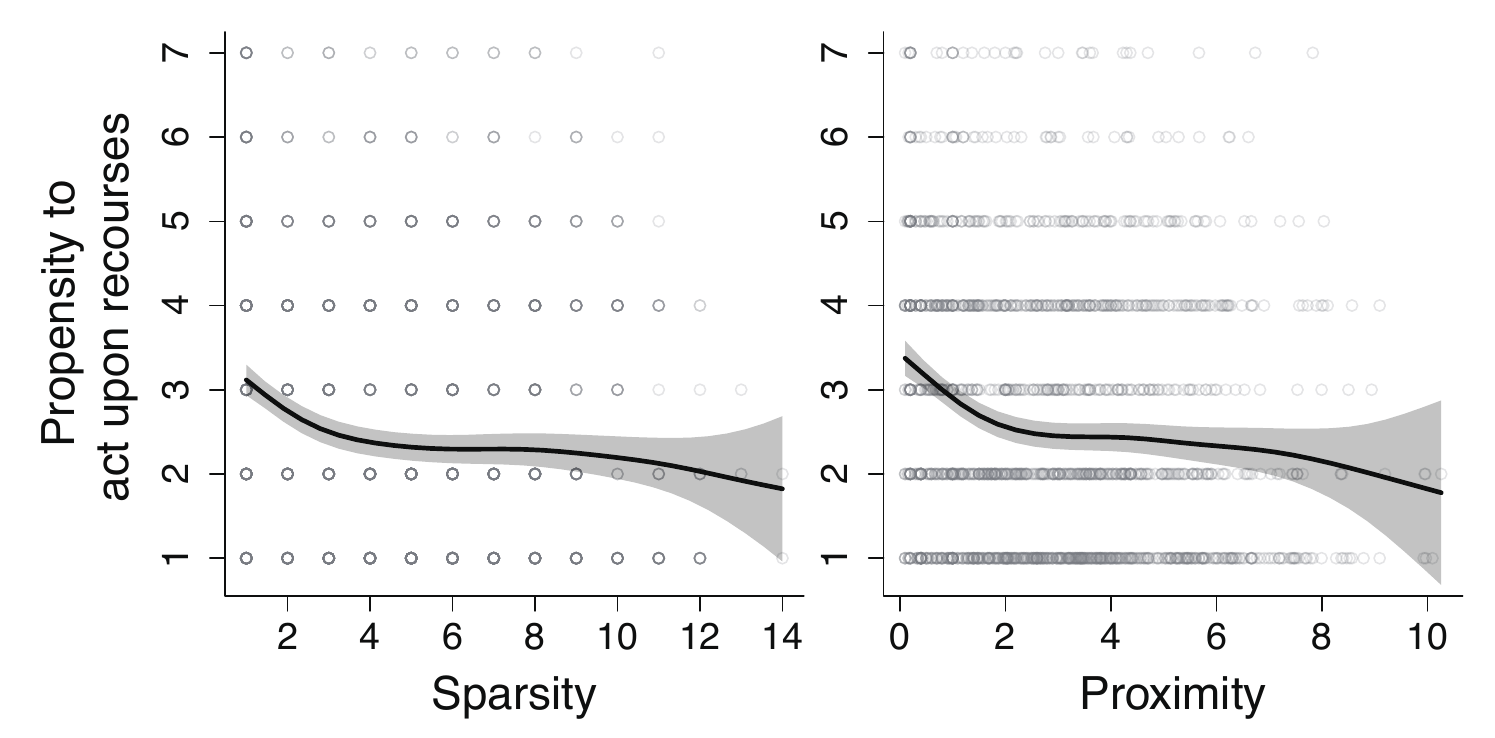}\label{subfig:act}
    }
    \caption{GAMM fits of the distance metrics to the propensity for acceptance (a) and action (b). The GAMMs include individual participant-specific effects as the nested random intercept and slopes. The error bars are 95\% confidence intervals.\label{fig:gamm}}
\end{figure*}
We formulated four distinct GAMMs based on combinations of the dependent variable (i.e., propensity to accept or to act upon recourses) and the explanatory variable (i.e., sparsity or proximity).
Figure~\ref{fig:gamm} illustrates the smoothing spline curves generated from applying these GAMMs to the experimental data. 
For a summary of statistics obtained from these models, please refer to Table S2 in the supplementary materials.

\subsection{Propensity to Accept Recourses}
\subsubsection{Quantitative Results}
We found that individuals' propensity to accept recourses is not associated with the distance metrics. 
As shown in the statistics from the GAMM detailed in Table S2, both sparsity and proximity did not exhibit a significant relationship with propensity for acceptance (Model-1: $F=0.00$, $p=1.00$; Model-2: $F=0.89$, $p=0.347$).
Figure~{\ref{subfig:acp}} illustrates the smoothing spline functions obtained from the GAMMs.
Propensity to accept recourses remains unchanged across different values of distance metrics.
This result challenges the foundational premise that sparser or more proximate recourses are more acceptable.

\subsubsection{Qualitative Results}
Analysis of the open-ended responses revealed that participants who deemed sparse or proximate recourses as unacceptable expressed skepticism about their failure to pass the car loan screening, especially when their profile data was almost identical to their counterfactual samples.
For example, P006 commented \textit{``Almost same''} when referring to a minor change (sparsity = 1; proximity = 0.167; Teleworking hours per day: 0-2 hours $\rightarrow$ 2-4 hours) and rated the recourse unacceptable (propensity for acceptance = 1). 
Similarly, P446 questioned \textit{``That's it?''} for a slight adjustment (sparsity = 1; proximity = 0.40; Educational background: Junior college $\rightarrow$ University) and gave a low score for accepting the recourse (propensity for acceptance = 2). 
P706 reported \textit{``Not particularly different compared to the current situation''} for a modification (sparsity = 2; proximity = 0.50; Position: Employee $\rightarrow$ Section Chief, Working hours per day: 10-12 hours $\rightarrow$ 8-10 hours) with a low acceptance rating (propensity for acceptance = 2).

Conversely, participants who rated non-sparse or non-proximate recourse acceptable often saw such challenging recourses as opportunities to introspect their capabilities and current status, attributing their loan application rejection to personal shortcomings.
For example, P567 stated \textit{``The career is too different''} for a significant shift (sparsity $=10$; proximity $=8.40$; e.g., Position: Employee $\rightarrow$ President), rating their acceptance high (propensity for acceptance $=7$). 
P739 described \textit{``I think it's a solid reason, so I don't blame it''} for a major transition (sparsity $=10$; proximity $=5.64$; e.g., Type of residence: Rental housing $\rightarrow$ Owned home, Service years: 3-5 years $\rightarrow$ 20- years), giving a high acceptance score (propensity for acceptance $=6$). 
Lastly, P732 admitted \textit{``Because I realize how much I'm not enough''} when faced with extensive changes (sparsity $=10$; proximity $=6.11$; e.g., Educational background: University $\rightarrow$ Graduate school (doctor), The number of side jobs: No $\rightarrow$ 2 jobs) and found it relatively acceptable (propensity for acceptance $=5$).

\subsection{Propensity to Act Upon Recourses}
\subsubsection{Quantitative Results}
The spline terms for sparsity and proximity exhibited significant associations with propensity for action, as seen in Table S2 (Model-3: $F=18.46$, $p<0.001$; Model-4: $F=20.81$, $p<0.001$).
In addition, the effective degrees of freedom (EDF) for the spline terms of sparsity and proximity were 4.25 and 4.32, respectively (Table S2). 
These results indicate a nonlinear rather than a strictly linear dependence between the propensity for action and the distance metrics. 
The smoothing spline functions of the distance metrics for the propensity for action are depicted in Figure~{\ref{subfig:act}}, revealing that the propensity to take action exhibits a linear and monotonic decrease in the range of small distance metric values, but remains relatively constant across other ranges.
In summary, the results indicate that participants were more willing to act on recourses that are both sparse and proximate, aligning with the assumption made in prior research. 
Notably, the propensity to act was highest at the minimal recourse distance but remained constant beyond a specific threshold.

\subsubsection{Qualitative Results}
Upon analyzing participants' explanations for their willingness to engage with sparse or proximate recourses, it became clear that they perceived these suggestions as readily executable tasks. 
For example, P674 stated \textit{``Because I think I can tackle it right now''} about a simple adjustment (sparsity = 1; proximity = 0.14; TOEIC: No $\rightarrow$ 0-400; propensity for action = 5), and P718 mentioned \textit{``If it was just working hours, I'd do it in a heartbeat''} regarding a minor change in working hours (sparsity = 1; proximity = 0.20; Working hours per day: 6-8 hours $\rightarrow$ 8-10 hours; propensity for action = 7).
Additionally, the perception that these recourses were straightforward to implement boosted participants' motivation to act. 
For example, P294 expressed  \textit{``Because the contents are easy to change''} when commenting on a simple transition (sparsity = 1; proximity = 0.20; The number of side jobs: 1 job $\rightarrow$ No; propensity for action = 7), and P655 described \textit{``All I have to do is wait a little''} regarding a minor increase in service years (sparsity = 1; proximity = 0.20; Service years: 0-1 years $\rightarrow$ 1-3 years; propensity for action = 6).

In contrast, when examining participant explanations on the assessment of non-sparse or non-proximate recourses that received low ratings of willingness to act, it was evident that participants cited increased burdens as a deterrent such as significant time and financial investment, a high volume of tasks to complete, or low perceived cost-effectiveness.
For example, P401 expressed \textit{``Because there's too much to do, it takes too much time and money''} when evaluating a major overhaul (sparsity = 11; proximity = 8.38; e.g., Residential prefecture: Other than Tokyo $\rightarrow$ Tokyo, Overseas working experience: No $\rightarrow$ Yes, LinkedIn: No $\rightarrow$ Yes; propensity for action = 2), P167 remarked, \textit{``There are too many items to clear''}, in response to extensive requirements (sparsity = 12; proximity =7.82; e.g., Type of residence: Rental housing $\rightarrow$ Owned house, Position: Supervisor $\rightarrow$ Section Head, TOEIC: 10-400 $\rightarrow$ 900-990; propensity for action = 1), and P468 stated \textit{``I don't want to change various things just to apply for a loan''}, highlighting the reluctance to undertake numerous changes (sparsity = 11; proximity = 9.09; e.g., Educational background: High school $\rightarrow$ Graduate school (doctor), Management career: No $\rightarrow$ 5-10 years, The number of side jobs: No $\rightarrow$ 1 job; propensity for action = 1).

\section{Discussion}
\paragraph{Implications.}
Our quantitative and qualitative results challenge the underlying assumption of algorithmic recourse, showing that the evaluation functions fail to capture individuals' propensity to accept recourses. 
Contrary to prior studies' assumption, we observed a nuanced dynamic: individuals expressed skepticism towards AI decisions when faced with recourses that are extremely ``close'' to their current situation, while they tended to acknowledge their inadequacies and accept AI decisions when presented with recourses suggesting substantial changes.
Furthermore, while the evaluation functions effectively capture the propensity for action up to a certain threshold, they fall short of representing it beyond that threshold.
To generate optimal recourses by maximizing individuals' willingness to act on recourses, setting firm upper limits of evaluation functions is crucial. 
Although this approach has been hinted at in prior works~\cite{VanLooveren2021,Pawelczyk2020}, our research validates its importance in enhancing user engagement with recourses.

\paragraph{Future directions.}
Designing evaluation functions that accurately reflect individuals' propensity to accept or act upon recourses remains a pressing challenge for advancing human-centered algorithmic recourse generation. 
The key to addressing this challenge lies in understanding the individual variance in recourse evaluation.
Our analysis using GAMMs highlights that individual subject-specific effects explain the propensity to accept or act upon recourses (see statistics of the variable \textit{uid} for all the models in Table S2).
This finding points to the importance of individual differences in shaping evaluation functions, which is overlooked in prior studies.
Future research should uncover key factors driving the individual differences and integrate them into evaluation functions.
By doing so, it would be possible to create evaluation functions that adapt to the nuanced acceptance/action propensity of each individual, enabling the generation of recourses tailored to each person's preferences and circumstances.

\paragraph{Limitations.}
Our experiment chose a car loan application scenario because financial screening is a paramount theme in recourse research.
However, further investigation is needed to ascertain if the observed results generalize across different contexts. 
Additionally, our participant pool was exclusively Japanese, highlighting the need for future studies to include more diverse demographics spanning various countries, languages, and cultures. 
This is especially critical in light of our discussions on individual differences, underscoring the importance of broadening the scope to ensure the applicability of our findings across diverse global populations.

\section{Conclusion}
Our empirical findings challenge the core assumption of algorithmic recourse generation research by demonstrating that the conventional evaluation functions fail to effectively reflect individuals' acceptance and action propensities. 
This insight underscores the pressing need for a paradigm shift in the design of evaluation functions for the next generation of human-centered algorithmic recourse AI. 
Specifically, it advocates for incorporating adaptive adjustments of evaluation functions to accommodate individual differences. 
Such a refined approach holds promise for delivering recourses more personalized and beneficial to individual users, significantly improving the efficacy of recourse generation.
We hope this research serves as a foundation for future scientific advancements and technological developments in XAI research.

\section*{Ethical Statement}
This study involving human subjects was reviewed and approved by the external ethics review committee, the Institutional Review Board of Public Health Research Foundation\footnote{\url{https://www.phrf.jp/rinri/} (only in Japanese)} (the protocol number: PHRF-IRB 23F0002).
All the procedures were performed in accordance with the guidelines.

\bibliographystyle{named}
\bibliography{ijcai}

\begin{appendix}
\setcounter{figure}{0}
\setcounter{table}{0}
\setcounter{footnote}{0}
\makeatletter
\renewcommand{\thefigure}{S\arabic{figure}}
\renewcommand{\thetable}{S\arabic{table}}
\renewcommand{\thefootnote}{\roman{footnote}}

\onecolumn

\begin{center}
    \textbf{\LARGE Supplementary materials}
\end{center}

\section{The Counterfactual Sample Pool for Recourse Construction}
\subsection{Overview}
To generate recourses tailored to participants' input data, we established a pool of counterfactual sample data in advance.
The overview of this data collection is as follows:
\begin{description}
    \item[Duration.] June 12th to 29th, 2023
    \item[Platform.] ASMARQ\footnote{ASMARQ Co., Ltd. A Japanese Marketing Research Company. \url{https://www.asmarq.co.jp/global/}}
    \item[Eligibility.] Company workers or government workers
    \item[Survey items.] See Section~\ref{ssubsec:survey}
    \item[Compensation.] 30 JPY 
\end{description}

Respondents provided their informed consent, confirming their understanding of the purpose and content of this survey, prior to their involvement. 
We pre-screened respondents for eligibility before obtaining responses to the survey items, and if they did not meet the eligibility criteria, we requested that they decline to participate.

As a result of this survey, we collected a total of 4,057 profile data entries. 
Subsequently, after excluding data from respondents who did not disclose their annual income, we retained a final pool of 3,683 profile data entries as a counterfactual sample pool.

In the scenario experiment described in the manuscript, we set a condition that participants should not have an annual income exceeding 10 million yen (Section 3.3, Participant Recruitment and Screening, (4)). 
This criterion was established to facilitate the provision of counterfactual samples to participants in the scenario experiment. 
For example, if someone with an annual income of over 10 million yen were to participate, we would need to prepare counterfactual samples with incomes of over 13 million yen, which is rare in our sample pool (only 1.47\% (54/3683) cases). 
Consequently, it becomes increasingly challenging to provide counterfactual samples for such participants. 
The purpose of this condition is to reduce this difficulty.

\subsection{Survey Items}~\label{ssubsec:survey}
\begin{description}
    \item[Instructions.] ``Please select the option that applies to you for each of the following items.''
    \item[Items and response options.] See Table~\ref{stab:survey_cfe_pool}. 
\end{description}
\begin{table}[h]
    \caption{Survey items and response options for collecting counterfactual sample pool.}
    \small
    \centering
    \begin{tabular}{lp{0.72\columnwidth}}
        \toprule
        Item & Option\\
        \midrule
        Annual income & 1. - 2M / 2. 2-3M / 3. 3-4M / 4. 4-5M / 5. 5-6M / 6. 6-7M / 7. 7-8M / 8. 8-9M / 9. 9-10M / 10. 10-11M / 11. 11-12M / 12. 12-13M / 13. 13-14M / 14. 14-15M / 15. 15-16M / 16. 16-17M / 17. 17-18M / 18. 18-19M / 19. 19-20M / 20. 20M - / 21. Choose not to disclose\\
        Residential preference & 1. Tokyo / 2. Other than Tokyo\\
        Type of residence & 1. Own house / 2. Rental housing\\
        Educational background & 1. High school / 2. Junior college / 3. University (bachelor) / 4. Graduate school (master) / 5. Graduate school (doctor)\\
        Workplace & 1. Private company / 2. Public institution\\
        Position & 1. Employee / 2. Supervisor / 3. Section Head / 4. Section Chief / 5. Assistant General Manager / 6. Manager / 7. General Manager / 8. Executive Director / 9. Senior Executive Director / 10. President\\
        Service years & 1. 0-1 year / 2. 1-3 years / 3. 3-5 years / 4. 5-10 years / 5. 10-20 years / 6. 20-years\\
        Management career & 1. No / 2. 0-1 year / 3. 1-3 years / 4. 3-5 years / 5. 5-10 years / 6. 10-20 years / 7. 20- years\\
        Working hours per day & 1. 0-2 hours / 2. 2-4 hours / 3. 4-6 hours / 4. 6-8 hours / 5. 8-10 hours / 6. 10-12 hours / 7. 12- hours\\
        Teleworking hours per day & 1. 0-2 hours / 2. 2-4 hours / 3. 4-6 hours / 4. 6-8 hours / 5. 8-10 hours / 6. 10-12 hours / 7. 12- hours\\
        The number of side jobs & 1. No / 2. 1 job / 3. 2 jobs / 4. 3 jobs / 5. 4 jobs / 6. 5- jobs\\
        Job change experience & 1. No / 2. Yes\\
        Overseas working experience & 1. No / 2. Yes\\
        Overseas study experience & 1. No / 2. Yes\\
        TOEIC Best score & 1. No / 2. 10-400 / 3. 400-495 / 4. 500-595 / 5. 600-695 / 6. 700-795 / 7. 800-895 / 8. 900-990\\
        Facebook app use & 1. No / 2. Yes\\
        LinkedIn app use & 1. No / 2. Yes\\
        \bottomrule
    \end{tabular}
    \label{stab:survey_cfe_pool}
\end{table}

\clearpage

\section{Scenario-based Questionnaires}
\subsection{Questionnaire for Screening Participants}
\begin{itemize}
    \item Please tell us your age.\\
    \lbrack \qquad \rbrack \ years old
    
    \item Please tell us your gender.
    1. Female\\
    2. Male\\
    3. Other
    
    \item Please tell us your occupation.\footnote{If a respondent selects \#6-\#11, she/he is excluded from the experiment.}\\
    1. Company worker (full-time)\\
    2. Company worker (contract)\\
    3. Company worker (temporary)\\
    4. Company worker (part-time)\\
    5. Government worker\\
    6. Self-employed/Freelance\\
    7. Homemaker\\
    8. Part-time job\\
    9. Student\\
    10. Unemployed/Retired\\
    11. Other

    \item Please tell us your annual income (JPY).\footnote{If a respondent selects \#10, she/he is excluded from the experiment. 1M = 1,000,000.}\\
    1. - 2M\\
    2. 2M - 3M\\
    3. 3M - 4M\\
    4. 4M - 5M\\
    5. 5M - 6M\\
    6. 6M - 7M\\
    7. 7M - 8M\\
    8. 8M - 9M\\
    9. 9M - 10M\\
    10. 10M -
    
    \item Please tell us about your thoughts regarding the purchase of a car.\footnote{If a respondent selects \#5, she/he is excluded from the experiment.}\\
    1. I plan to purchase within a year.\\
    2. I plan to purchase within three years.\\
    3. I plan to purchase within five years.\\
    4. While the timing is undecided, I plan to purchase.\\
    5. I do not plan to purchase.

    \item Please select all of the following that apply to your current loan situation.\footnote{If a respondent selects \#1-\#7, she/he is excluded from the experiment.}\\
    1. Card loans\\
    2. Automobile loans\\
    3. Mortgage loans\\
    4. Education loans\\
    5. Free loans\\
    6. Business loans\\
    7. Other loans\\
    8. I do not have any loans
\end{itemize}

\clearpage

\subsection{Questionnaire for Profile Data Submission}
Please respond to the following questions, imagining the scenario below.\\

\noindent
Currently, you are considering taking out a two-year car loan to purchase a car equivalent to one-third of your annual income.
\begin{center}
    \includegraphics[height=5cm]{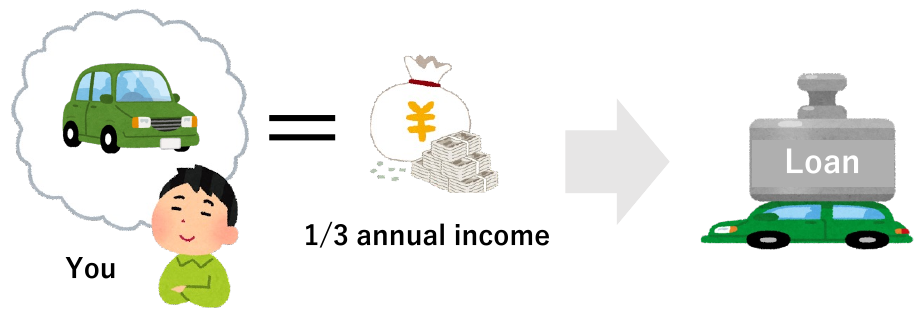}
\end{center}

\noindent
Right now, you are at a financial institution undergoing a loan assessment. In this assessment, an AI system determines the approval or denial of the loan based on your profile data.
\begin{center}
    \includegraphics[height=5cm]{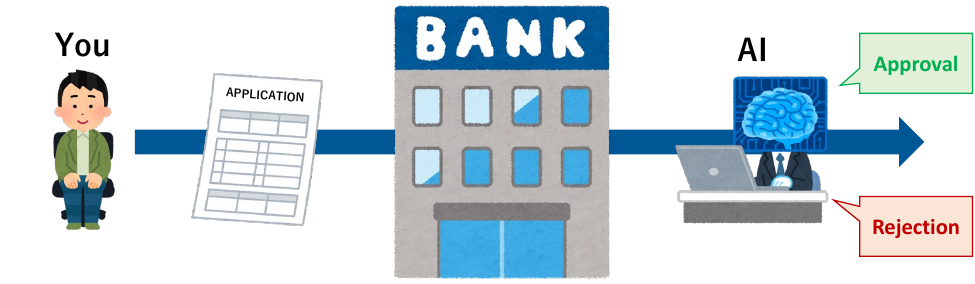}
\end{center}

\noindent
To undergo this evaluation, you have decided to submit your own profile data.
\begin{center}
    \includegraphics[height=5cm]{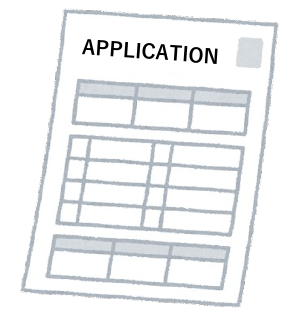}
\end{center}

\hfill (Go to the next page)

\clearpage

\noindent
Please select the option that applies to you for each of the following items.\\

\begin{small}
    \centering
    \begin{tabular}{lp{0.72\columnwidth}}
        \toprule
        Item & Option\\
        \midrule
        Residential preference & 1. Tokyo / 2. Other than Tokyo\\
        Type of residence & 1. Own house / 2. Rental housing\\
        Educational background & 1. High school / 2. Junior college / 3. University (bachelor) / 4. Graduate school (master) / 5. Graduate school (doctor)\\
        Workplace & 1. Private company / 2. Public institution\\
        Position & 1. Employee / 2. Supervisor / 3. Section Head / 4. Section Chief / 5. Assistant General Manager / 6. Manager / 7. General Manager / 8. Executive Director / 9. Senior Executive Director / 10. President\\
        Service years & 1. 0-1 year / 2. 1-3 years / 3. 3-5 years / 4. 5-10 years / 5. 10-20 years / 6. 20-years\\
        Management career & 1. No / 2. 0-1 year / 3. 1-3 years / 4. 3-5 years / 5. 5-10 years / 6. 10-20 years / 7. 20- years\\
        Working hours per day & 1. 0-2 hours / 2. 2-4 hours / 3. 4-6 hours / 4. 6-8 hours / 5. 8-10 hours / 6. 10-12 hours / 7. 12- hours\\
        Teleworking hours per day & 1. 0-2 hours / 2. 2-4 hours / 3. 4-6 hours / 4. 6-8 hours / 5. 8-10 hours / 6. 10-12 hours / 7. 12- hours\\
        The number of side jobs & 1. No / 2. 1 job / 3. 2 jobs / 4. 3 jobs / 5. 4 jobs / 6. 5- jobs\\
        Job change experience & 1. No / 2. Yes\\
        Overseas working experience & 1. No / 2. Yes\\
        Overseas study experience & 1. No / 2. Yes\\
        TOEIC Best score & 1. No / 2. 10-400 / 3. 400-495 / 4. 500-595 / 5. 600-695 / 6. 700-795 / 7. 800-895 / 8. 900-990\\
        Facebook app use & 1. No / 2. Yes\\
        LinkedIn app use & 1. No / 2. Yes\\
        \bottomrule
    \end{tabular}
\end{small}
\\ \\

\noindent
The results of the assessment will be communicated to you in 2 weeks. 
At that time, we will inquire about your impressions of the assessment results. Thank you for your cooperation.

\clearpage

\subsection{Questionnaire for Recourse Evaluation}
You had come for a two-year loan assessment with the intention of purchasing a car equivalent to one-third of your annual income. 
But unfortunately, you were declined in the loan assessment.
\begin{center}
    \includegraphics[height=5cm]{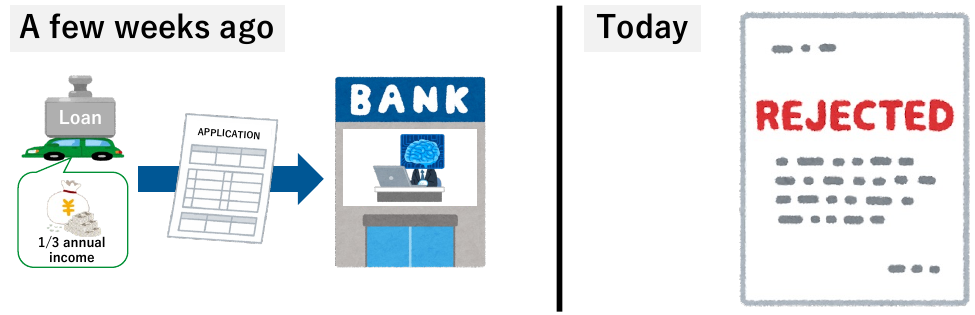}
\end{center}

\noindent
The AI system used in this assessment can provide a failed applicant with an ideal profile for her/his current profile, like the figure below, as a plan of action to ensure that her/his next application will be approved.
\begin{center}
    \includegraphics[width=\columnwidth]{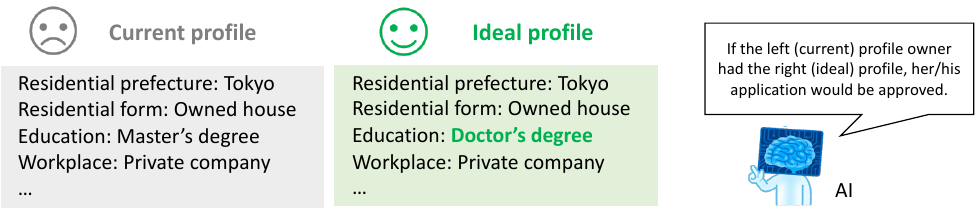}
\end{center}

\noindent
So, you decided to consider several action plans.

\hfill (Go to the next page)

\clearpage

\noindent
Here is the 1st plan\footnote{In the experiment, participants evaluated five different recourses (i.e., 1st to 5th plans). All questions for each were the same.}, created by AI. 
If you had the profile shown on the right, your application would have been approved\footnote{For protecting the privacy of participants, profile data items not recommended for modification are masked in gray in this figure.}. 
The AI is proposing action plans with minimal change costs to reduce the burden on applicants as much as possible.
\begin{center}
    \includegraphics[width=0.55\columnwidth]{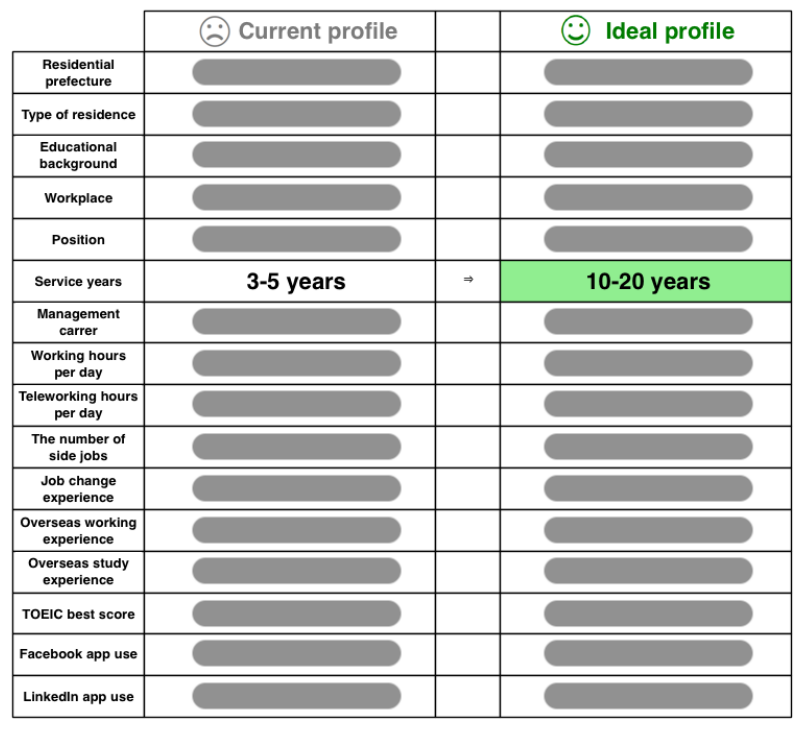}
\end{center}

\begin{itemize}
    \item Is the AI system's plan a reasonable explanation for the rejection of your loan application?\\
    1. Strongly no\\
    2. No\\
    3. Slightly no\\
    4. Neutral\\
    5. Slightly yes\\
    6. Yes\\
    7. Strongly yes
    
    \item For the above question, why did you make the evaluation you did?\\
    \lbrack \qquad \rbrack

    \item Is it impossible for you to implement each change item presented by the AI system?\footnote{Only the items that a recourse suggested changing appeared on the actual survey form as response items. For example, if a recourse suggested changing "Educational background" (e.g., from "University (bachelor)" to "Graduate school (doctor)") and "TOEIC Best score" (e.g., from "600-695" to "800-895"), items \#3 and \#14 are displayed on the survey form and the others are not.}\\
    1. Residential prefecture: Yes / No\\
    2. Type of residence: Yes / No \\
    3. Educational background: Yes / No\\
    4. Workplace: Yes / No\\
    5. Position: Yes / No\\
    6. Service years: Yes / No\\
    7. Management career: Yes / No\\
    8. Working hours per day: Yes / No\\
    9. Teleworking hours per day: Yes / No\\
    10. The number of side jobs: Yes / No\\
    11. Job change experience: Yes / No\\
    12. Overseas working experience: Yes / No\\
    13. Overseas study experience: Yes / No\\
    14. TOEIC Best score: Yes / No\\
    15. Facebook app use: Yes / No\\
    16. LinkedIn app use: Yes / No

    \item Would you carry out the plan presented by the AI system to obtain loan approval?\\
    1. Strongly no\\
    2. No\\
    3. Slightly no\\
    4. Neutral\\
    5. Slightly yes\\
    6. Yes\\
    7. Strongly yes

    \item For the above question, why did you make the evaluation you did?\\
    \lbrack \qquad \rbrack 
\end{itemize}

\clearpage

\section{Statistics from Generalized Additive Mixed Models}
We applied four generalized additive mixed models to the experimental data and derived the following statistical outcomes.

\begin{table}[h]
    \centering
    \begin{threeparttable}
        \caption{GAMM results for propensity for acceptance and action fitted by sparsity or proximity accounting for individual participant-specific effects. In the GAMMs, the distance metrics are introduced as smoothing function variables, the participant IDs (uid) are included as random intercept (r.i.) and slopes (r.s.) of linear predictors for coefficients of the smoothing functions, and the intercept (Intercept) is modeled as fixed effects.}
        \small
        \begin{tabular}{cclcrrrrrrr}
            \toprule
            \multicolumn{1}{c}{Model} & \multicolumn{1}{c}{Outcome} & \multicolumn{1}{c}{Variable} & \multicolumn{1}{c}{Smth} & \multicolumn{1}{c}{Coef.} & \multicolumn{1}{c}{S.E.} & \multicolumn{1}{c}{$t$ value} & \multicolumn{1}{c}{EDF} & \multicolumn{1}{c}{$F$ value} & \multicolumn{1}{c}{$p$ value} & \multicolumn{1}{c}{AIC}\\        
            \midrule
            \multirow{4}{*}{Model-1} & \multirow{4}{*}{Propensity for acceptance} & Intercept & & 3.28 & 0.07 & 45.60 & & & $<$0.001 & \multirow{4}{*}{5597.44}\\ 
             & & Sparsity & \checkmark & & & & 1.01 & 0.00 & 1.000\\
             & & uid (r.i.) & \checkmark & & & & 235.35 & 9.21 & $<$0.001\\
             & & uid (r.s.) & \checkmark & & & & 142.98 & 5.21 & $<$0.001\\
            \midrule
            \multirow{4}{*}{Model-2} & \multirow{4}{*}{Propensity for acceptance} & Intercept & & 3.29 & 0.07 & 45.93 & & & $<$0.001 & \multirow{4}{*}{5590.47}\\ 
             & & Proximity & \checkmark & & & & 1.00 & 0.89 & 0.347\\
             & & uid (r.i.) & \checkmark & & & & 258.69 & 9.40 & $<$0.001\\
             & & uid (r.s.) & \checkmark & & & & 133.20 & 4.00 & $<$0.001\\
            \midrule
            \multirow{4}{*}{Model-3} & \multirow{4}{*}{Propensity for action} & Intercept & & 2.63 & 0.07 & 38.06 & & & $<$0.001 & \multirow{4}{*}{5261.35}\\ 
             & & Sparsity & \checkmark & & & & 4.25 & 18.46 & $<$0.001\\
             & & uid (r.i.) & \checkmark & & & & 269.32 & 10.37 & $<$0.001\\
             & & uid (r.s.) & \checkmark & & & & 99.44 & 2.89 & $<$0.001\\
            \midrule
            \multirow{4}{*}{Model-4} & \multirow{4}{*}{Propensity for action} & Intercept & & 2.63 & 0.07 & 38.77 & & & $<$0.001 & \multirow{4}{*}{5280.33}\\ 
             & & Proximity & \checkmark & & & & 4.32 & 20.81 & $<$0.001\\
             & & uid (r.i.) & \checkmark & & & & 288.58 & 8.49 & $<$0.001\\
             & & uid (r.s.) & \checkmark & & & & 69.98 & 1.13 & 0.009\\
            \bottomrule
        \end{tabular}
        \label{tab:my_label}
    \end{threeparttable}
    \begin{tablenotes}
        \small
        \item Notations.
        \item uid: Participant IDs.
        \item Smth: Whether or not the variable is a smoothing term.
        \item Coef, Std. Err., $t$ value: Coefficient values, standard errors, and $t$ values of the variables introduced as fixed effects.
        \item EDF: Effective degrees of freedom of the smoothing term. Higher EDF implies more complex, wiggly splines. When a variable’s EDF is close to 1, it is close to being a linear term.
        \item $F$ value: The statistic value from the F-test for a smoothing term to confirm whether it is equal to zero for all coefficients that constitute a single spline term. If it is significant, the smoothing term is an influential variable. 
        \item $p$ value: $p$ value of F-test.
        \item AIC: Akaike information criteria.
    \end{tablenotes}
\end{table}

\section{Code and Data Availability}
The experimental data generated and analyzed by the authors during this study contain or relate to sensitive and personal records and so are governed by the regulations of the authors' institution. 
Thus they are not publicly available, but they can be accessed from the corresponding author upon reasonable request, along with the source code necessary for replicating the statistical analysis, results, and figures and tables, within the bounds of the authors' institutional regulations.

\end{appendix}

\end{document}